# Enabling Inclusive Systematic Reviews: Incorporating Preprint Articles with Large Language Model-Driven Evaluations

Rui Yang[1†], Jiayi Tong[2†], Haoyuan Wang[3], Hui Huang[4], Ziyang Hu[5], Peiyu Li[3], Nan Liu[1], Christopher J. Lindsell[3], Michael J. Pencina[3], Yong Chen[6*], Chuan Hong[3,7*]

[1] Centre for Quantitative Medicine, Duke-NUS Medical School, Singapore, Singapore

[2] Department of Biostatistics, Johns Hopkins Bloomberg School of Public Health, Baltimore, MD, USA

[3] Department of Biostatistics and Bioinformatics, Duke School of Medicine, Durham, NC, USA

[4] Children's Healthcare of Atlanta, Emory University School of Medicine, Atlanta, GA, USA

[5] Department of Genome Sciences, University of Virginia, Charlottesville, VA, USA

[6] Department of Biostatistics, Epidemiology and Informatics, Perelman School of Medicine, University of Pennsylvania, Philadelphia, PA, USA

[7] Duke Clinical Research Institute, Durham, NC, USA

†Rui Yang and Jiayi Tong are equally contributed

*Yong Chen and Chuan Hong are equally contributed

*Correspondence: Chuan Hong

Email: chuan.hong@duke.edu Phone: 903-526-9514

## ABSTRACT

### Objectives

Systematic reviews in comparative effectiveness research require timely evidence synthesis. With the rapid advancement of medical research, preprint articles play an increasingly important role in accelerating knowledge dissemination. However, as preprint articles are not peer-reviewed before publication, their quality varies significantly, posing challenges for evidence inclusion in systematic reviews.

### Materials and Methods

We developed *AutoConfidence* (automated confidence assessment), an advanced framework for predicting preprint publication, which reduces reliance on manual curation and expands the range of predictors, including three key advancements: (1) automated data extraction using natural language processing techniques, (2) semantic embeddings of titles and abstracts, and (3) large language model (LLM)-driven evaluation scores. Additionally, we employed two prediction models: a random forest classifier for binary outcome and a survival cure model that predicts both binary outcome and publication risk over time.

### Results

The random forest classifier achieved an area under the receiver operating characteristic curve (AUROC) of 0.747 using all features. The survival cure model achieved an AUROC of 0.731 for binary outcome prediction and a concordance index of 0.667 for time-to-publication risk.

## Discussion

Our study advances the framework for preprint publication prediction through automated data extraction and multiple feature integration. By combining semantic embeddings with LLM-driven evaluations, *AutoConfidence* significantly enhances predictive performance while reducing manual annotation burden.

## Conclusion

*AutoConfidence* has the potential to facilitate incorporation of preprint articles during the appraisal phase of systematic reviews, supporting researchers in more effective utilization of preprint resources.

## INTRODUCTION

Comparative effectiveness research (CER) is a pivotal component in the landscape of health care, focusing on evaluating the effectiveness of different treatments and interventions across various health conditions.[1] The essence of CER lies in its ability to guide clinical decisions through evidence-based insights.[1] In this context, timely and comprehensive systematic reviews are paramount, as they collate and compare the latest research, providing critical insights for clinical practice and policy formulation.[2]

The continuous evolution of medical treatments and interventions, particularly the ongoing advancements in managing chronic conditions such as cardiovascular diseases (CVDs), highlights the importance of staying abreast of the latest research to ensure effective and evidence-based patient care.[3] As one of the leading chronic diseases worldwide, CVDs account for over 20 million deaths annually, and this burden continues to rise with global population aging.[4,5] At the same time, CVD-related research has expanded rapidly, with new clinical trials and observational studies emerging at an accelerating pace.[6] Catalyzed by the COVID-19 pandemic, the rapid dissemination of findings through preprint platforms has become increasingly common across medical fields,[7] highlighting the need for more inclusive evidence synthesis in high-stakes areas like CVD.

Over the last decade, preprint databases such as medRxiv have become instrumental in the swift dissemination of new findings, allowing real-time tracking of research progress and fostering the development of immediate solutions.[8] Preprint articles provide early access to cutting-edge research,[9] and have increasingly become a key component of the evolving evidence ecosystem in medicine. Given this shift, incorporating both preprint articles and peer-reviewed publications into systematic reviews represents a forward-looking approach.[2] Such integration can substantially enhance the comprehensiveness and timeliness of evidence synthesis in comparative effectiveness research (CER), ensuring that clinical decisions are informed by the most current and expansive data available.

Importantly, a previous study on the medRxiv database showed that 77% of preprint articles are successfully published within 24 months, and over 80% of these articles maintain consistency with the final published journal articles in terms of sample size, primary endpoint results, and overall interpretation.[8] Although the inclusion of preprint articles in systematic reviews broadens the scope of available data, it presents significant challenges due to their preliminary nature and variable quality prior to formal peer review.[10,11] The urgency in finding effective treatments and interventions for diseases, coupled with the reliance on rapidly emerging literature, increases the risk of incorporating studies with lower methodological quality, which could potentially misinform clinical guidelines and harm patients.[10,11] Therefore, incorporating preprint articles into systematic reviews requires a careful balance to maintain scientific rigor while capitalizing on their timely insights.

Previously, Tong et al. addressed the challenge of integrating preprint articles into systematic reviews through the introduction of the concept of a "confidence score", derived from a survival cure model to predict the likelihood that preprint articles will eventually be published in peer-reviewed journals.[2] However, this approach relies on the manual extraction of a limited set of predictors, which restricts its scalability and robustness in capturing factors influencing preprint publication.

Recent advancements in natural language processing (NLP) and large language models (LLMs) have significantly enhanced the process of automated feature extraction and evaluation of the quality of preprint articles.[12–17] These technologies can efficiently identify and extract key metadata elements such as author affiliations, study type, and usage metrics of preprint articles, significantly streamlining the data collection process for systematic reviews.[13] Additionally, text embedding techniques enable the generation of contextual representations of preprint articles, capturing the semantic information within the text.[12,18] Meanwhile, state-of-the-art LLMs can be leveraged to generate comprehensive quality assessments of preprint articles, evaluating aspects such as originality, significance, quality of presentation, depth of research, and interest to readers.[14] The integration of the above multidimensional approaches, including automated metadata extraction, semantic embeddings, and LLM-driven evaluations, has the potential to enable a more nuanced and comprehensive assessment of the textual quality and scientific relevance of preprint articles, thereby improving the feasibility of integrating preprints into evidence synthesis.

Extending prior research on the “confidence score”, we propose *AutoConfidence*, which is designed to advance the framework by incorporating automated data extraction, semantic embeddings and LLM-driven evaluations. By reducing reliance on manual curation and including a broader set of predictors, our approach introduces key advancements: (1) scalable automated data extraction from preprint articles through NLP techniques, (2) leveraging text embeddings of titles and abstracts to capture semantic information, and (3) incorporating LLM-driven evaluations, relying on their advanced capabilities in understanding textual content. *AutoConfidence* could facilitate more inclusive and timely evidence synthesis, addressing the long-standing challenge of integrating high-quality yet unpublished research into systematic reviews. Specifically, by applying *AutoConfidence* during the appraisal phase, researchers can, on the one hand, evaluate the publication likelihood of preprint articles and, on the other hand, use the LLM-driven scores it provides to assess their quality, thus providing quantitative indicators of their potential contribution to evidence synthesis. As a demonstration, we applied *AutoConfidence* to research on CVDs, a domain where rapid access to emerging evidence is critical for informing clinical guidelines and decision-making.

## MATERIALS AND METHODS

In this section, we introduce the *AutoConfidence* framework for predicting preprint publication. Figure 1 provides an overview. First, we present the automated data extraction process (Figure 1a). Then, we elaborate on the approaches to generating predictive features through semantic embeddings (Figure 1b) and LLM-driven evaluations (Figure 1c). Finally, we employ two models to capture different aspects of the prediction task (Figure 1d): a random forest classifier provides a binary classification within a certain time range, while a survival cure model predicts both binary outcome and publication risk over time, offering a more comprehensive understanding of the publication process.

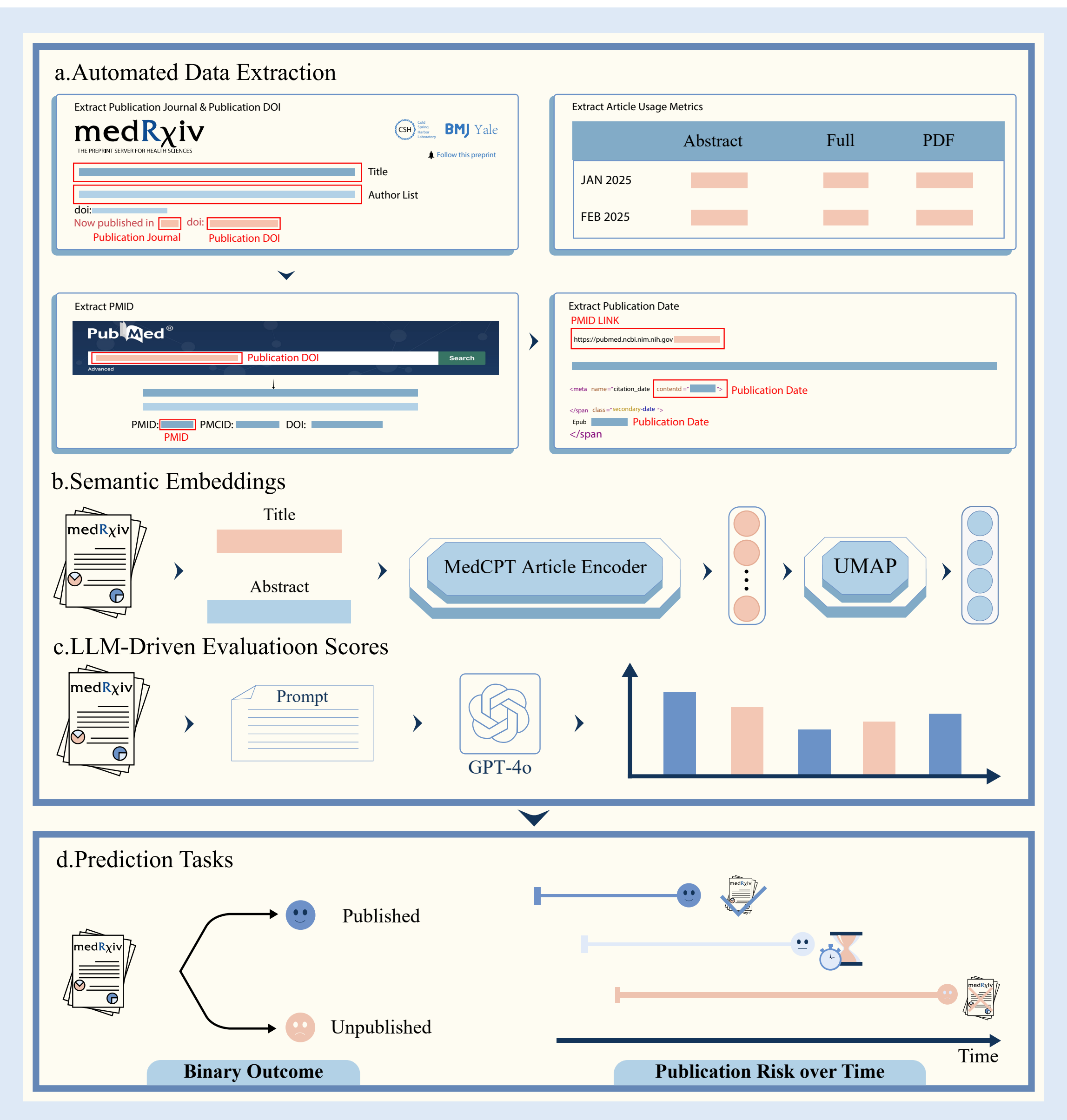

**Figure 1. Overview of the Framework for Preprint Publication Prediction.** The framework consists of four components: (a) Automated data extraction pipeline that extracted related information from medRxiv (for all preprint articles) and PubMed (for published articles); (b) Semantic embeddings of the title and abstract; (c) LLM-

driven evaluation scores; (d) Two prediction tasks - binary outcome and publication risk over time.

**Automated Data Extraction**

We chose medRxiv, a preprint database which focuses on biomedical and clinical research, making it a suitable source for our study. We utilized the "medrxivr" R package to retrieve metadata from medRxiv.[19] This package not only provides programmatic access to the Cold Spring Harbor Laboratory API for downloading preprint metadata such as titles, abstracts, and author lists, but also supports access to a static snapshot of the medRxiv repository. To identify CVDs-related preprint articles, we used search terms including "cardiovascular disease", "heart disease", "hypertension", "stroke", "atherosclerosis", "heart attack", "vascular disease", "blood pressure", and "atrial fibrillation". These search terms were chosen in consultation with domain experts to ensure comprehensive coverage of relevant studies.

Building on this metadata collection, we developed an automated pipeline to track publication status of preprint articles. By parsing the source code of medRxiv webpages, our pipeline monitored real-time publication status and extracted key details, including the publication digital object identifier (DOI), the publication journal for published articles, and article usage metrics (the access counts for abstract, full text, and pdf) for all preprint articles. Subsequently, we mapped the extracted publication DOIs to the PubMed database to obtain precise publication dates (prioritizing electronic publication dates when available), providing essential temporal data for the survival cure model. In addition, to evaluate the quality of preprint articles that have been published, we leveraged the Journal Citation Reports (JCR) from Web of Science[20] to obtain journal evaluation metrics, including journal quartile rankings and impact factors. For published articles, we selected those published in Q1 journals (top 25%) with an impact factor ≥5, which helped to differentiate article quality to some extent.

## Predictive Features

### Semantic Embeddings

To effectively capture the semantic features of preprint articles, we utilized the MedCPT (bioMedical Contrastive Pre-trained Transformers) model, specifically designed for biomedical literature.[21] MedCPT, which was trained on 255 million query-article pairs from PubMed search logs, demonstrates exceptional performance in capturing medical terminology and textual semantics. We employed the "MedCPT Article Encoder" to generate vector representations of the titles and abstracts of preprint articles. Given the potential redundancy in high-dimensional embedding vectors, we applied the UMAP (Uniform Manifold Approximation and Projection) algorithm to reduce the embedding dimensionality to four dimensions, a choice determined by sensitivity analysis and practical considerations, ensuring key semantic information while lowering computational complexity.[22] This nonlinear dimensionality reduction algorithm preserves both local and global structures of the data, enabling efficient feature representation for subsequent visualization and predictive analyses.

### LLM-Driven Evaluations

To enrich the quality assessment metrics for preprint articles, we introduced an LLM-driven evaluation framework, providing an innovative quantitative approach to their assessment. Specifically, we employed OpenAI's GPT-4o model (*version 2024-11-20*) to mimic the peer review process of medical preprint articles.[23] The evaluation framework includes five dimensions: Originality, Significance, Quality of Presentation, Depth of Research, and Interest to Readers, each scored on a scale of 1 to 10. In our context, Originality refers to the novelty of the research question; Significance assesses the potential impact of the research question; Quality of Presentation focuses on the article's structure and clarity; Depth of Research reflects the comprehensiveness of the analysis and critical evaluation; and Interest to Readers considers the practical value of the article, including its guidance and applicability for reference in real-world settings.

Meanwhile, to ensure evaluation reliability and consistency, we designed the detailed scoring prompt and provided high-quality and low-quality reference instances as benchmarks, ensuring the LLM's scoring aligned with expectations in medical research (the evaluation prompt can be found in Figure S1 of the Supplementary Figures). The LLM analyzed each article's full text and produced structured outputs, including scores across these five dimensions. These multidimensional scores served as new predictive features, providing additional content quality metrics for the survival cure model.

In addition to the evaluation scores, we leveraged the LLM to directly predict whether a preprint article would be accepted for publication, outputting a binary outcome, and compared its prediction accuracy with other models (the prompt can be found in Figure S2 of the Supplementary Figures).

## Predictive Models

### Random Forest for Binary Outcome

We employed a random forest classifier, implemented using Python's "scikit-learn" package[24], to predict the binary outcome of whether a preprint article posted by September 1, 2023, was published by January 4, 2025. The classifier utilized a comprehensive set of predictors, including the above-mentioned semantic embeddings generated by the MedCPT model, multidimensional LLM-driven evaluation scores, and usage metrics from the first three months after the preprint article was submitted. To ensure data reliability, we excluded preprint articles that were posted on medRxiv and subsequently published within the first three months, as they were likely near acceptance when posted and did not accurately reflect the initial state of the preprint articles. The remaining labeled data were used to train the classifier.

### Survival Cure Model for Time-to-Publication

Meanwhile, we applied a survival cure model to predict the likelihood of preprint publication. In the survival cure model, the publication dynamics of preprint articles are divided into two components: the cure component and the survival component. The cure component estimates the probability that a preprint article will ultimately be published, modeling a binary outcome: 1 if susceptible (i.e., may eventually be published) vs. 0 if cured (i.e., never published). The survival component, modeled using a Cox proportional hazards model,[25] focuses only on the susceptible group (preprint articles with non-zero probability of being published), and models the time-dependent publication risk (i.e., the likelihood of being published at each point in time, given that the preprint article has not yet been published).

Specifically, the model expresses the probability that a preprint article remains unpublished at time $t$ using the following mixture survival function:

$$S_T(t|x) = \pi(x)S(t|x) + \big(1 - \pi(x)\big) \quad (1)$$

where $\pi(x)$ represents the probability that a preprint article will ultimately be published, with $x$ being the covariates including LLM-driven evaluation scores and semantic embeddings, and $S(t|x)$ denotes the survival function for the preprint article at risk of being published at time $t$. The analysis applied an administrative censoring date of January 4, 2025 (the cutoff date after which we no longer tracked publication status), to define whether a preprint article could be classified as unpublished or published. The probability in the cure component was modeled with logistic regression as follows:

$$\pi(x) = \frac{exp(x^{\top}\gamma)}{1 + exp(x^{\top}\gamma)} \quad (2)$$

where $\gamma$ represents the regression coefficient vector corresponding to the covariates $x$.

For preprint articles at risk of being published, their time to publication is modeled using a Cox proportional hazards model, with the survival function given as:

$$S(t|x) = S_0(t)^{exp(x^\top \beta)} \quad (3)$$

where $S_0(t)$ is the baseline survival function, and $\beta$ represents the regression coefficient vector for the covariates $x$. The corresponding hazard function is:

$$h(t|x) = h_0(t)exp(x^\top \beta) \quad (4)$$

where $h_0(t)$ is the baseline hazard function.

This two-component structure enables the survival cure model to simultaneously estimate the probability of non-publication (Equation 2) and the time-dependent risk of publication (Equation 4), providing a comprehensive view of publication dynamics up to the cutoff date.

**Evaluations**

We evaluated the models' performance using five-fold cross-validation to ensure robustness and generalizability.[26] We used the area under the receiver operating characteristic curve (AUROC) and concordance index (C-index) to evaluate the performance of the binary outcome and publication risk over time, respectively. Additionally, we performed 100 bootstrap iterations to calculate confidence intervals (CI).[27]

## RESULTS

The search covered all preprint articles posted on medRxiv before September 1, 2023, initially yielding 2,162 articles. We then filtered the articles based on their publication status: for published articles, only those with an impact factor ≥5 and published in Q1 journals (top 25%) were retained. Additionally, we excluded articles with missing data (without journal quartile information) and removed duplicate entries. Ultimately, 698 unpublished preprint articles and 385 published articles were included for subsequent analysis.

**Comparative Analysis of Published and Unpublished Preprint Articles**

As shown in Figure 2(A), we compared GPT-4o's scores for unpublished preprint articles and published articles across five dimensions, and the independent samples t-test consistently showed significant differences ($p < 0.001$) for all dimensions. In terms of originality, published articles received an average score of 6.33, compared to 5.47 for unpublished preprint articles. A similar difference was observed in the significance dimension, with published articles scoring 6.95 and unpublished preprint articles scoring 6.01. Quality of presentation received relatively lower scores, with published articles scoring 5.23 and unpublished preprint articles scoring 4.70. The most notable difference was seen in the depth of research dimension, where published articles scored 7.42, compared to 6.31 for unpublished preprint articles—a gap of 1.11 points. For interest to readers, published articles scored 5.99, while unpublished preprint articles scored 5.11. In addition, the scoring patterns indicate that dimensions such as depth of research and significance were rated relatively higher, whereas quality of presentation and interest to readers received more conservative scores. We further analyzed the distribution of semantic embeddings generated by MedCPT, which were reduced to four dimensions using UMAP. As shown in Figure 2(B), we compared the embedding distributions of unpublished and published preprint articles across each dimension. The density plots reveal varying degrees of distributional differences between the two groups, with noticeable shifts in certain dimensions. This indicates that semantic embeddings can capture intrinsic differences in the textual content of unpublished and published articles.

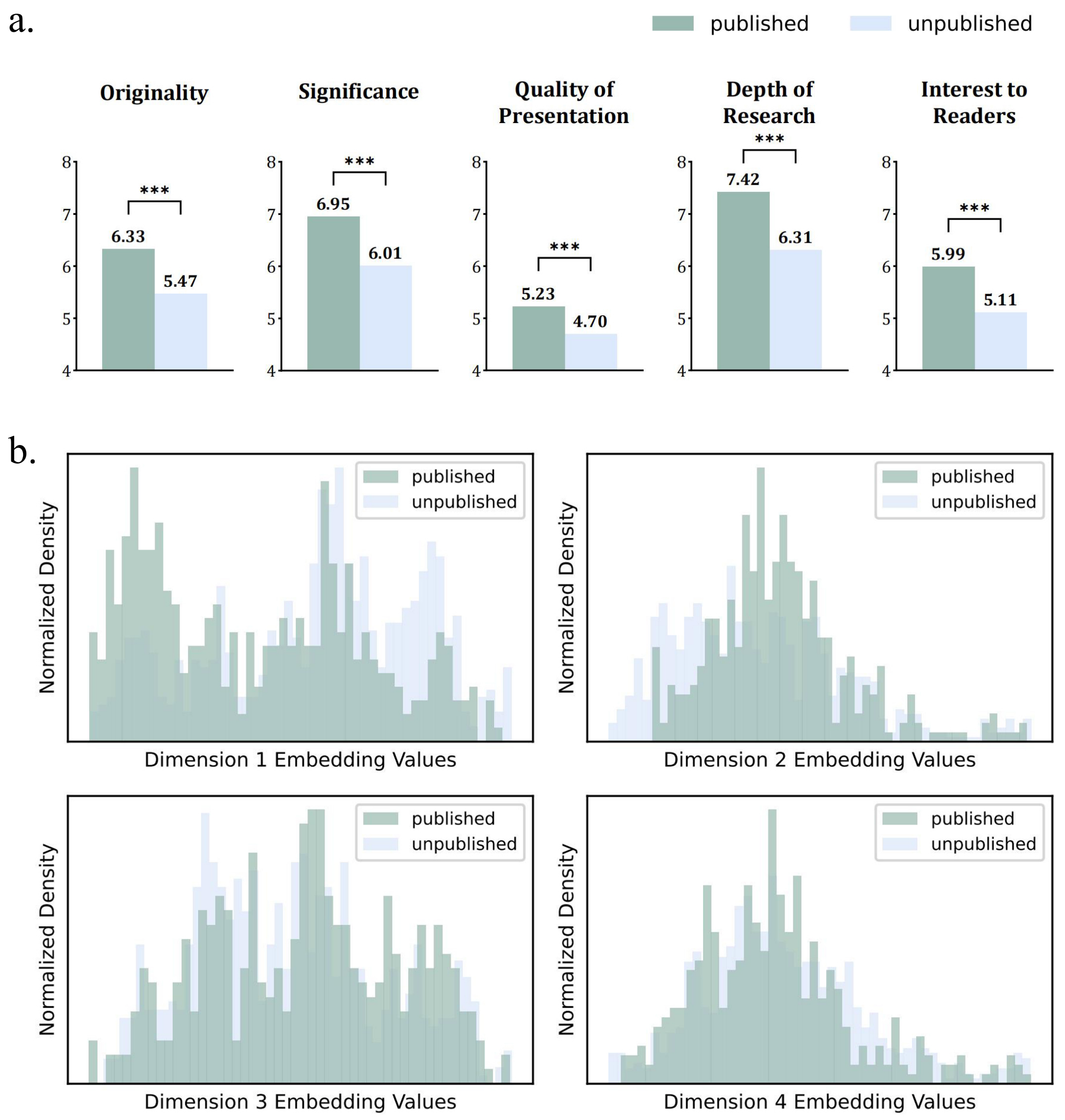

**Figure 2. Comparative Analysis of Published and Unpublished Preprint Articles.**

**(a) GPT-4o Evaluations for Published and Unpublished Preprint Articles Across Five Dimensions.** Figure (2a) presents the average GPT-4o evaluation scores for published articles (green) and unpublished preprint articles (blue) across five dimensions: Originality, Significance, Quality of Presentation, Depth of Research, and Interest to Readers. Published articles consistently received higher scores across all dimensions. The triple asterisks (***, $p < 0.001$) indicate statistically significant differences between the two groups. **(b) Semantic Embeddings for Published and Unpublished Preprint Articles.** Figure (2b) illustrates the distribution differences of semantic embeddings, with green representing published articles and blue representing unpublished articles.

**Evaluation of Random Forest for Binary Outcome**

To ensure data reliability, we excluded 44 preprint articles that were published within three months of submission. For the remaining data, the binary outcome is defined as the publication status of each preprint article, categorized as "published" or "unpublished." As shown in Figure 3, the model using only LLM-driven evaluation scores achieved an AUROC of 0.692 (95% CI: 0.663-0.729). Incorporating semantic embeddings improved performance, yielding an AUROC of 0.733 (95% CI: 0.702-0.760). Further adding article usage metrics from the first three months led to the highest performance, with an AUROC of 0.747 (95% CI: 0.716-0.774).

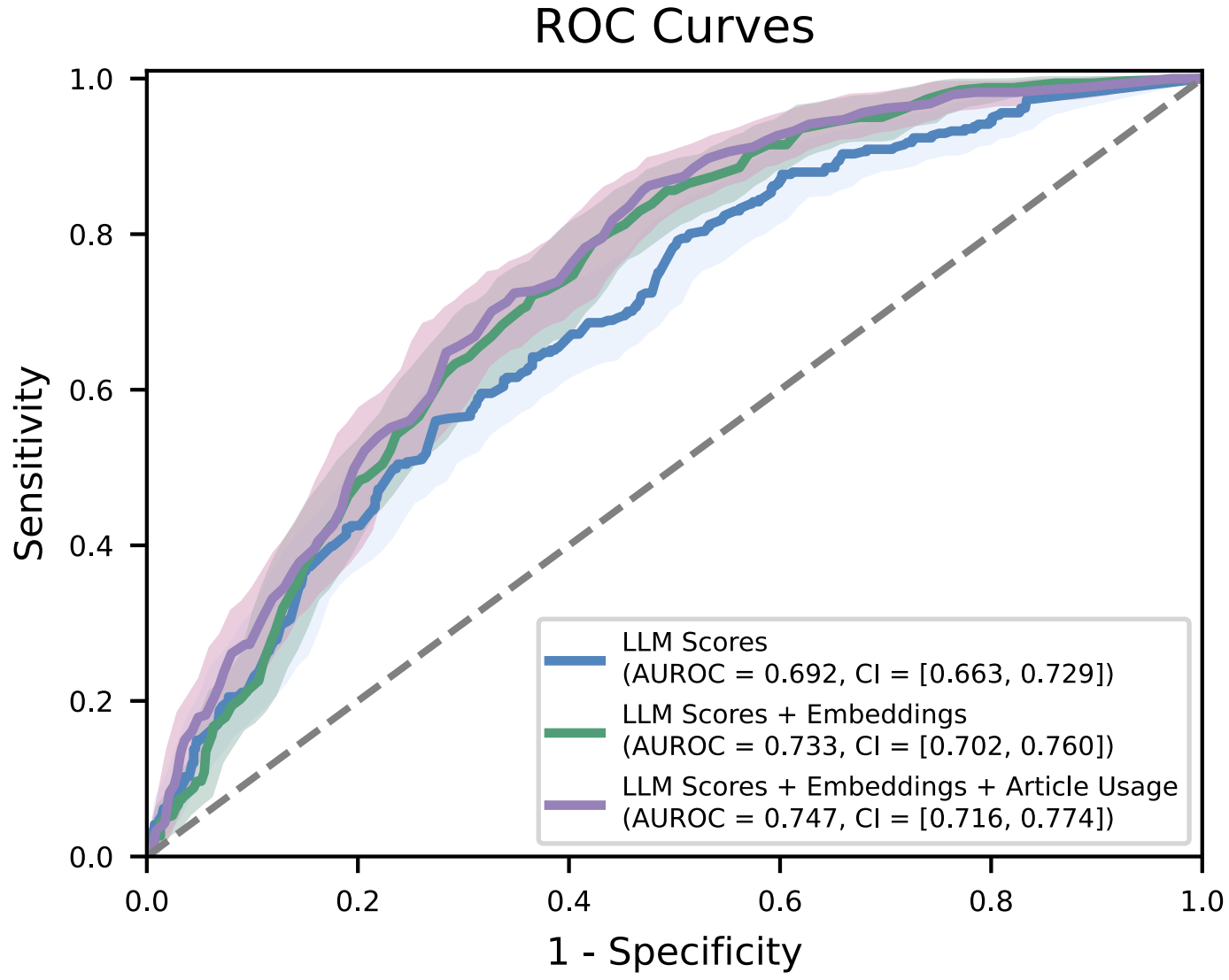

**Figure 3. ROC Curves for the Random Forest Classifier.** The figure shows the ROC curves for the random forest model predicting preprint publication. The model performance was evaluated using three different feature combinations: (1) LLM-driven evaluation scores only (blue curve): AUROC = 0.692 (95% CI: 0.663-0.729); (2) LLM-driven evaluation scores + semantic embeddings (green curve): AUROC = 0.733 (95% CI: 0.702-0.760); (3) LLM-driven evaluation scores + semantic embeddings + article usage (red curve): AUROC = 0.747 (95% CI: 0.716-0.774).

**Evaluation of the Survival Cure Model**

To evaluate the performance of the survival cure model, we evaluated both its binary outcome prediction and publication risk over time, as shown in Figure 4. For binary outcomes, the model achieved an AUROC of 0.716 (95% CI: 0.680-0.746) using LLM-driven evaluation scores, improving to 0.731 (95% CI: 0.697-0.760) after incorporating semantic embeddings. For publication risk over time, the model achieved a C-index of 0.660 (95% CI: 0.633-0.687) with LLM-driven evaluation scores, and incorporating semantic embeddings further improved the C-index to 0.671 (95% CI: 0.646-0.700). These results indicate that semantic embeddings enhance the model's performance for both prediction tasks. In addition, since the survival cure model already incorporates temporal modeling, we did not include time-varying article usage metrics.

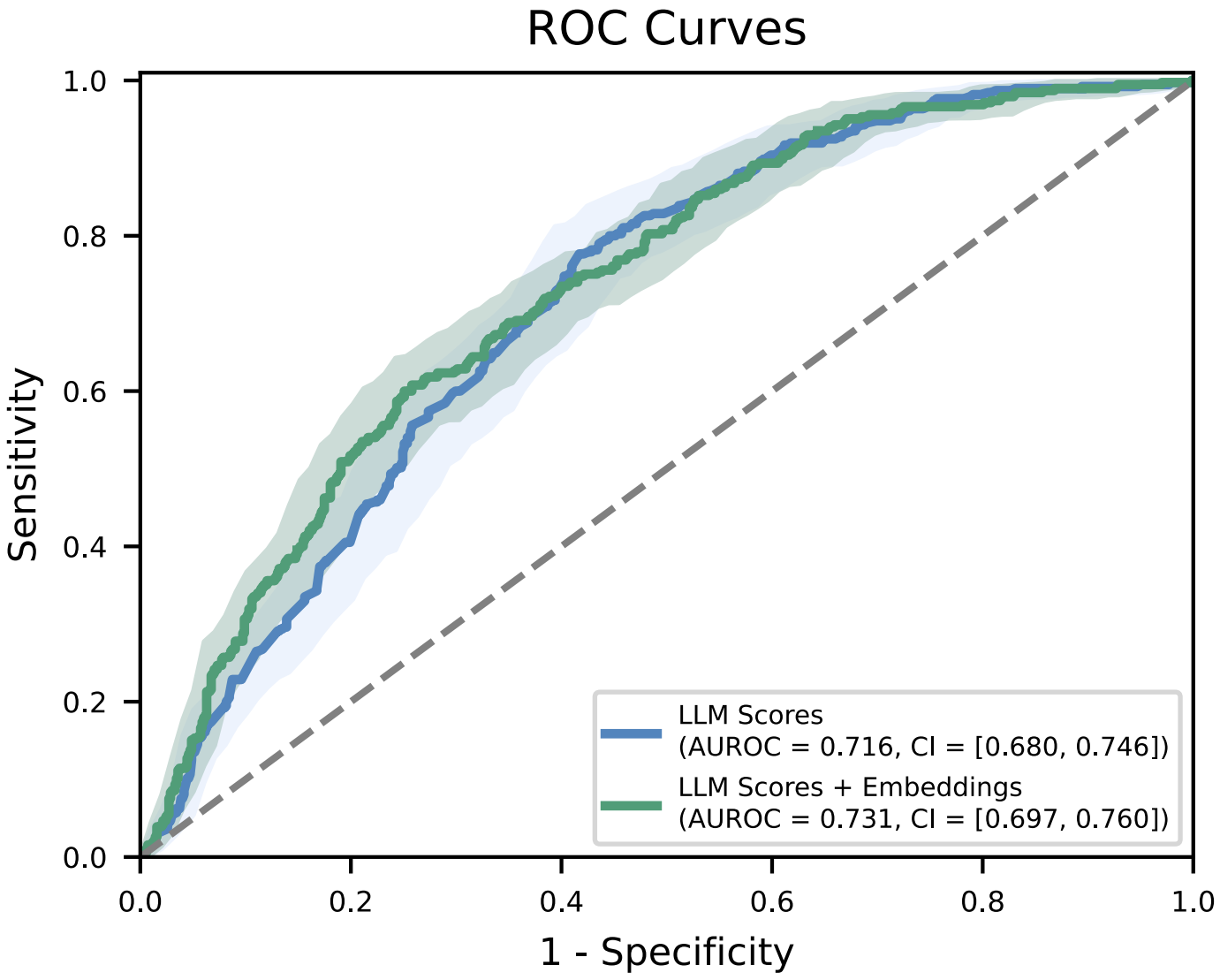

**Figure 4. ROC Curves for the Survival Cure Model.** The figure shows the ROC curves for the cure component of the survival cure model. The model performance was evaluated using two different feature combinations: (1) Metadata + LLM-driven evaluation scores only (blue curve): AUROC = 0.716 (95% CI: 0.680-0.746); (2) Metadata + LLM-driven evaluation scores + semantic embeddings (green curve): AUROC = 0.731 (95% CI: 0.697-0.760).

### Direct Binary Prediction Using LLM

To evaluate the potential of LLMs for direct prediction, we assessed GPT-4o's performance in binary classification of preprint publication outcomes. By directly outputting predictions (0 for unpublished, 1 for published), GPT-4o achieved an accuracy of 0.683. This performance is comparable to models using only LLM-driven evaluation scores as predictive features (with a classification threshold of 0.5, the random forest achieved 0.676, and the survival cure model achieved 0.662). When semantic embeddings were incorporated, the performance of these models improved further (random forest: 0.694, survival cure model: 0.698). These results suggest that while LLM provides a strong baseline for direct prediction, traditional models leveraging a combination of multiple features can achieve higher accuracy.

| | Accuracy | | |
|---|---|---|---|
| | LLM-Driven Predictions/Evaluations | with Semantic Embeddings | with Article Usage |
| **GPT-4o** | 0.683 | - | - |
| **Survival Cure Model** | 0.662 | 0.698 | - |
| **Random Forest** | 0.676 | 0.694 | 0.693 |

**Table 1. Binary Classification Accuracy.** The accuracy demonstrated the comparative performance between GPT-4o's direct binary prediction and two models (random forest classifier and survival cure model). A dash (-) indicates that the corresponding feature combination was not evaluated.

## DISCUSSION

This study developed *AutoConfidence,* a framework for predicting preprint publication, integrating automated data extraction, semantic embeddings, and LLM-driven evaluations. We evaluated the framework using 1,083 preprint articles related to CVDs (698 unpublished and 385 published in high-impact journals), employing both the random forest classifier for binary classification and the survival cure model for estimating both publication likelihood and publication risk over time.

Our evaluation reveals the effectiveness of combining multiple features in both prediction tasks. The random forest classifier demonstrated progressive improvement: using only LLM-driven evaluation scores achieved an AUROC of 0.692, incorporating semantic embeddings increased it to 0.733, and adding three-month article usage metrics further improved it to 0.747. These results suggest that multidimensional features have the potential to enhance the predictive capability, with semantic embeddings capturing textual information and article usage metrics providing early indicators of impact. For the survival cure model, in binary outcome prediction, it achieved an AUROC of 0.716 with LLM-driven evaluation scores, improving to 0.731 with semantic embeddings. In predicting publication risk over time, it showed a C-index of 0.658, while incorporating semantic embeddings improved performance to 0.667, indicating that semantic features could provide complementary information.

In the broader context of evidence synthesis and systematic review workflows, various semi-automated/automated toolkits have been developed, such as Rayyan[28] for collaborative literature screening, RobotReviewer[29] for specifically assessing risk of bias in published randomized controlled trials, DistillerSR[30] for comprehensive systematic review management, and GRADEpro GDT[31] for evidence quality grading. However, these toolkits primarily focus on processing peer-reviewed publications, with limited support for quality assessment and integration of preprint articles.

The primary contribution of this study lay in advancing the use of NLP techniques for automatically extracting data from preprint articles, reducing reliance on manual annotation, and improving the scalability of data processing. Additionally, we explored the role of semantic embeddings and LLM-driven evaluations in predicting preprint publication. This framework, particularly through its integration of LLM-driven evaluations, has the potential to open new avenues for systematic incorporation of preprint articles in evidence-based medicine. As preprint articles play an increasingly important role in academic communication, automated evaluation approaches could help researchers more effectively utilize preprint resources. By combining LLM capabilities with traditional metrics, our framework provides a more comprehensive and scalable solution for evaluating the quality of preprint articles, potentially facilitating more timely and thorough systematic reviews.

In practical applications, *AutoConfidence* is primarily used during the appraisal phase of systematic reviews, and it generates two types of outputs that can be flexibly applied depending on the review's objectives. Initially, the survival cure model provides a confidence score representing the estimated probability that a given preprint article will eventually be published in a peer-reviewed journal. This score can be used either as a binary decision threshold (e.g., include only preprint articles with a predicted publication probability ≥70%, where the threshold can be selected based on the desired precision or recall) or as a continuous appraisal weight to inform sensitivity analyses or evidence grading. For instance, reviewers may choose to include all preprint articles meeting general eligibility criteria while evaluating whether the results change if low-confidence preprints are excluded or downweighted. In addition, *AutoConfidence* provides LLM-driven evaluation scores across five dimensions (originality, significance, depth of research, quality of presentation, and interest to readers), with scores ranging from 1 to 10. These scores can help reviewers prioritize which preprints to pursue for full-text review, particularly when there is a large number of eligible preprint articles. They can also be used to inform qualitative evidence grading frameworks or to complement existing risk-of-bias assessments.

However, several limitations should be noted. First, our analysis was confined to preprint articles in the CVD domain, and the generalizability of the framework to other medical domains remains untested. Second, our use of strict journal criteria (Q1 quartile and impact factor ≥5) to define high-quality publications may be overly conservative, as important studies can also appear in specialized journals with lower impact factors. Notably, this journal filtering was used only during model training and does not reflect how *AutoConfidence* is intended to be used in systematic reviews. However, we acknowledge that journal-based metrics are imperfect proxies for study quality and may reinforce publication bias. Lastly, the LLM-driven evaluation exhibited distinct scoring patterns across different dimensions, which may reflect both the inherent characteristics of preprint articles and limitations of the current LLM-driven evaluation method, warranting further validation.

Future work should validate the generalizability of *AutoConfidence* across broader medical domains, consider more inclusive criteria for defining "high-quality publication", and extend its application to other forms of gray literature to better reflect the diversity of academic publishing. Meanwhile, further exploration could be conducted on how to directly incorporate semantic embeddings and structured features such as article usage data into LLM inputs to evaluate their ability to distinguish heterogeneous information.

## CONCLUSION

In conclusion, *AutoConfidence* facilitates the integration of preprint articles into the appraisal phase of systematic reviews through an innovative framework that combines automated data extraction, semantic embeddings, and LLM-driven evaluations. By improving the performance of publication prediction, it enables more timely and comprehensive evidence synthesis while reducing the burden of manual curation.

# Supplementary Figures

**LLM-Driven Evaluation Scoring Prompt**

**INSTRUCTION**

You are a highly stringent peer reviewer for a cardiology medical journal. Please rate the following article on a scale of 1 to 10 based on originality, significance, quality of expression, depth of research, and interest to readers. Your ratings need to differentiate the quality of these articles as much as possible; do not give similar scores to each article.

**Rating 9:**
{good_benchmark}

**Rating 2:**
{bad_ benchmark}

**INPUT**

{article}

**OUTPUT**

Only return the result in the following format. Don't generate anything else.

```
{{
   "Originality":,
   "Significance:,
   "Quality of Presentation":,
   "Depth of Research":,
   "Interest to Readers":,
}}
```

**Figure S1. LLM-Driven Evaluation Scoring Prompt.** The prompt is designed for GPT-4o to evaluate preprint articles across five dimensions (Originality, Significance, Quality of Presentation, Depth of Research, and Interest to Readers). The prompt includes scoring guidelines and reference examples to ensure consistent evaluation.

**LLM-Driven Preprint Publication Prediction Prompt**

**INSTRUCTION**

You are an expert peer reviewer specializing in cardiology. Your task is to evaluate whether the provided article indicates potential for publication in Q1-Ranked (Top 25%) cardiovascular journals. Meanwhile, we provide the corresponding institution information. If the article meets the standards for publication in Q1-Ranked (Top 25%) cardiovascular journals, return 1; otherwise, return 0. Additionally, I have provided two simplified benchmark instances for reference - one that was published and one that was not published.

**Instance (Published in Q1 Journal):**
{published_instance}

**Instance (Not Published in Q1 Journal):**
{unpublished_instance}

**INPUT**

{article}

**OUTPUT**

Return the result in the following format.

{{
  “Publish”: <0 or 1>.
}}

**Figure S2. Binary Publication Prediction Prompt.** The prompt is designed for GPT-4o to directly predict whether a preprint article would be published. The prompt instructs the model to output a binary prediction (0 for unpublished, 1 for published) based on the article's content.